\documentclass[sigconf]{acmart}

\AtBeginDocument{%
  \providecommand\BibTeX{{%
    \normalfont B\kern-0.5em{\scshape i\kern-0.25em b}\kern-0.8em\TeX}}}

\setcopyright{acmcopyright}
\copyrightyear{2018}
\acmYear{2018}
\acmDOI{10.1145/1122445.1122456}

\acmConference[Woodstock '18]{Woodstock '18: ACM Symposium on Neural
  Gaze Detection}{June 03--05, 2018}{Woodstock, NY}
\acmBooktitle{Woodstock '18: ACM Symposium on Neural Gaze Detection,
  June 03--05, 2018, Woodstock, NY}
\acmPrice{15.00}
\acmISBN{978-1-4503-9999-9/18/06}
\usepackage{xcolor}

\begin{document}

\title{Predicting Future Opioid Incidences Today}

\author{Sandipan Choudhuri, Kaustav Basu, Kevin Thomas, Arunabha Sen}
\affiliation{%
  \institution{NetXT Lab, School of Computing, Informatics and Decision Systems Engineering}
  \streetaddress{Arizona State University}
  \city{Arizona State University, Tempe}
  \state{AZ}
}

\renewcommand{\shortauthors}{Choudhuri and Basu, et al.}

\begin{abstract}
According to the Center of Disease Control (CDC), the Opioid epidemic has claimed more than 72,000 lives in the US in 2017 alone. In spite of various efforts at the local, state and federal level, the impact of the epidemic is becoming progressively worse, as evidenced by the fact that the number of Opioid related deaths increased by 12.5\% between 2016 and 2017. Predictive analytics can play an important role in combating the epidemic by providing decision making tools to stakeholders at multiple levels - from health care professionals to policy makers to first responders. Generating Opioid incidence heat maps from past data, aid these stakeholders to visualize the profound impact of the Opioid epidemic. Such post-fact creation of the heat map provides \textit{only} retrospective information, and as a result, may not be as useful for preventive action in the current or future time-frames. In this paper, we present a novel deep neural architecture, which learns subtle spatio-temporal variations in Opioid incidences data and accurately predicts \textit{future} heat maps. We evaluated the efficacy of our model on two open source datasets- (i) The Cincinnati Heroin Overdose dataset, and (ii) Connecticut Drug Related Death Dataset.

\end{abstract}

%



\maketitle

\section{Introduction}

Opioids are drugs, prescribed by health care professionals to relieve patients from pain and often lead to addiction. According to the CDC, the Opioid epidemic has claimed more than 72,000 lives in US in 2017 alone, up by 12.5\% from 2016, which led the US government to declare the epidemic, as a public health emergency in October 2017. Blue Cross Blue Shield \cite{BCBS} stated in their 2017 report that 21\% of their commercially insured members filled at least one Opioid prescription in 2015. Their data show that members with an Opioid Use Disorder (OUD) diagnosis, grew to 493\% over a seven year period, from 2010 to 2016. 


Owing to this growing threat, researchers in the medical and analytical domains are looking at ways by which, careful analysis of relevant data, may provide some useful insights into the epidemic. Relevant data may comprise of prescription patterns of health care professionals (such as a doctor of medicine, dentist, nurse practitioner, etc.), Opioid consumption patterns of patients, time and locations of Opioid related incidences, etc. The insights obtained after analysis of such data, can be taken into consideration while formulating response at multiple levels - from health care professionals to policy makers to first responders. Although a few health insurance companies and data analytics firms have examined this important issue, analytical research findings from the analysis of publicly available Opioid data are sparse.  

Heat maps (hot spots) of opioid incidences are created by governmental and non-governmental agencies, in an effort to visualize the impact of the epidemic. Oftentimes, these maps are created using past data. Policy makers, law enforcement agencies, etc. analyze these heat maps to gain insights into the spread of the epidemic, over a geographical area. Resource allocation decisions, such as establishment of new Medication Assisted Treatment (MAT) centers, stocking up on Naloxone doses, organizing rehabilitation programs etc. are often based on the analysis of Opioid incidence hot spots, obtained from previous data. It may be noted that such post-fact generation of heat maps provides respective authorities with only retrospective information. It may not be as useful for preventive action, in the current or subsequent time-frame. It will be of benefit to these professionals, if they are provided with the analytical tools to \textit{predict} the heat map for the upcoming time-frame (week, month, year, etc.), by analyzing historical heat maps.

In this work-in-progress paper, we have developed a novel deep neural architecture for predicting the Opioid heat maps for the future. Our model takes past daily heat maps as input, and predicts the heat map for the subsequent day. We have tested the efficacy of our model on two publicly available Opioid incidence datasets, (i) The Cincinnati Heroin Overdose dataset, and (ii) Connecticut Drug Related Death dataset. To summarize, our contributions in this paper are as follows:

\begin{itemize}
	\item To the best of our knowledge, this is the first known work where machine-learning tools have been utilized to accurately predict future Opioid incidence heat maps. 
	\item Our spatio-temporal architecture has been designed to accurately predict the heat maps, even when there are missing data in between the time frames taken into consideration. Moreover, our training approach is robust to scenarios where the time period of data availability is considerably short.  
\end{itemize}



\section{Related Work}

\label{sec:RW}
The analyses of the effects of Opioids on the population has been thoroughly studied in medical domain. Authors in \cite{chou2015effectiveness}, evaluated evidences on the effectiveness and harms of long-term Opioid therapy for chronic pain in adults. The examination of the association of maximum prescribed daily Opioid dose and dosing schedule, with risk of Opioid overdose death among patients with cancer, chronic pain, acute pain, and substance use disorders, was studied in \cite{bohnert2011association}. Authors in \cite{cicero2017prescription}, performed a systematic literature review, using a qualitative approach to examine the development of an Opioid-use disorder from the point of initial exposure. In the analytical domain, data science researchers from IBM Research \cite{IBM} and experts at Watson Health have recently embarked on applying data analytics and machine learning techniques to combat the Opioid problem \cite{IBM2}. Authors in \cite{mackey2017twitter} studied illegal sales of prescription Opioids online, utilizing Twitter data. Authors in \cite{rice2012model} have developed a model to identify patients at risk for prescription Opioid abuse, using drug claims data. The use of machine learning techniques for surveillance of drug overdose was studied in \cite{neill2018machine}. The application of deep neural networks, such as recurrent neural networks, for classifying patients on Opioid use was illustrated in \cite{che2017deep}. Authors in \cite{acion2017use} highlighted the use of machine learning and deep learning for predicting substance use disorder treatment success.

The problem of capturing underlying patterns in time sequences has been a long standing problem in the field of Computer Vision. Recently, generating future patterns have been studied by various research groups \cite{isola2017image},\cite{johnson2016perceptual}, \cite{nguyen2017plug}, \cite{zhang2017stackgan}. Authors in \cite{zhang2017deep} have developed spatio-temporal residual networks for crowd flow prediction. Predictive analysis of opioid incidences involves drawing inferences from a large set of features, many of which are difficult to identify and procure. In order to circumvent this overhead, we propose a methodology to predict the future hot spots (heat maps) by looking at hot spots of the previous months. The future hot spot prediction task requires deep understanding of the trajectory of the previous incidence locations. We extend the concept presented by Srivastava et. al. \cite{srivastava2015unsupervised} to capture this property by modeling our framework on an encoder-decoder architecture, consisting of time-distributed convolutional layers. We transform the given task into a supervised-learning problem by sequencing monthly opioid-incidence heat maps into fixed-length spatio-temporal representations and utilize them to predict heap maps for the subsequent months.



\section{Datasets}
\label{sec:D}
To predict the heat maps of the future, we first collected data from two sources - the Cincinnati Heroin Overdose dataset and Connecticut Drug Related Death Dataset.

\subsection{Cincinnati Heroin Overdose Dataset}

The Cincinnati dataset \cite{Cincinnati}, $CD$, has been launched by the City of Cincinnati, and contains detailed information regarding an Opioid related incident in Cincinnati, such as location (latitudinal and longitudinal coordinates), time, EMS response type, neighborhood, etc. that require an EMS dispatch (45 features in total). This dataset contains incidences ranging from January 2016 till the present day. As of May 21 2019, there are 7191 recorded Opioid incidences spanning 1235 days, spread across the neighborhoods of the city. 

\subsection{Connecticut Drug Related Death Dataset}

The Connecticut Accidental Drug (Opioid) Related Death \cite{Connecticut}, $CN$, dataset is very similar to $CD$.  Every row in $CN$ denotes an Opioid related incidence (death in this case), and contains 32 features for each death - sex, race, age, city/county of residence, city/county of death, latitude/longitudinal information, etc. This dataset is not updated as regularly as $CD$, but has 4083 mortality records. 
\section{Approach}

\label{sec:PF}


In this section, we formalize the problem of predicting accurate heat maps of the future. To generate the heat map for a particular region over a specified time-frame, we utilize the latitudinal and longitudinal information present in the datasets. We have developed a model which predicts the heat map for the subsequent day, having observed (or learnt from) the past data. In the following, we discuss in detail, our procedure for the generation of the incidence heat maps for the Cincinnati dataset only, as the procedure is replicated in a similar manner for the Connecticut dataset. 

We construct $CD'$ $(CD'\subset CD)$, by extracting latitudinal and longitudinal information. A tuple $CD'_{\{d,x,y\}}$ contains three entries, where $d \in \{1,..,1235\}$, denotes the day and $\{x, y\}$ represents the latitudinal and longitudinal coordinates respectively. For each value of $d$, a gray-scale image $HM_d$ (heat map for day $d$) is generated by plotting the latitudinal and longitudinal coordinates $\{x,y\}$. $HM_d$ has intensities ranging between $[0,255]$, with 255 in locations where maximum incidences have occurred and 0 in locations with no incidence data. All plots have been plotted on a predefined scale space, scaled to the Cincinnati land area. 

The heat maps generated thus far \textit{only} highlight the \textit{exact locations} of the Opioid incidences and \textit{ignores} the spatial influence of the geographic region (for e.g. large open parks, abandoned buildings, under bridges, etc.). Therefore, capturing incidence locations and the underlying spatial influences are crucial to the problem at hand. We manage to tackle this issue by modeling the spread of influence, by fitting a Gaussian distribution $G(\mu, \sigma)$ over $HM_{d}$ (with $\sigma$ controlling he spread of influence), thereby yielding a smoothed variant $HM'_{d}$. Thus, having generated the set of $HM'_d$ images, we can formally state our problem.


\textbf{Problem:} \textit{Given a set of heat map images $HM'_d$ of $n$ consecutive days, the objective is to predict the heat map for the subsequent day}. Due to resource limitations, we could only consider $n$ (where $2 \leq n \leq 8$) consecutive days, for capturing the spatial and temporal dependencies of Opioid incidences.


The proposed task of predicting future hot spots can be formulated as a supervised-learning problem. More formally, a training data-label pair is represented as a stacked volume of the $n$ heat maps $\{HM'_{t},HM'_{t+1},...,HM'_{t+n-1}\}$ corresponding to images of $n$ consecutive \textit{days} as input, and the $HM'_{t+n}$th image as the train label, where, $1\leq t\leq$ s and $s \leq |HM'|-n-1$. Here, $s$ controls the train-test split and is based on the cardinality of $HM'$. The value of $s$ is based on the respective dataset, and is explained in the subsequent paragraphs. Given missing data, i.e., if there are no incidences reported for a day, the corresponding heatmap is a blank image of zero values.

For testing, given a stack $<HM'_{t},HM'_{t+1},..., HM'_{t+n-1}>$ of $n$ consecutive heat-maps, the model will output the heat map $HM'_{t+n}$ for the $(t+n)^{th}$ day. Here, $(s+1)\leq t\leq |HM'| - n$. As mentioned earlier, to capture the dependency of incidence counts on varying scales of daily information, we worked with large values of $n$ ( $2 \leq n \leq 8$), and found $n = 6$ to be optimal.

In this work, propose a novel deep neural network to solve the above mentioned problem. The proposed learning task can be modeled under a Generative Adversarial Learning framework that handles spatio-temporal data. Consequently, we construct a model comprising of Attention-Based Stacked Convolutional LSTMs as the generative model $G$ to predict heatmap for the next time-frame. The discriminative model $D$ is based on the CNN architecture and performs convolution operations on the input heatmap in order to estimate the probability whether a sequence comes from the dataset or is produced by $G$. We use adversarial loss to train the combined model ($G$ and $D$). The prime intuition behind using this loss is that it can simulate the operating zones of incidences through historically available indicator data.

However, in practice, minimizing adversarial loss alone cannot guarantee satisfying the predictions. $G$ can generate samples that can confuse $D$ without even being close to the actual distribution of future heatmaps. In order to tackle this problem, we propose a prediction error loss that minimized the L1-distance between the actual and generated samples. The model is trained using the a joint loss function formed by the combination of adversarial and prediction error losses.

\section{Experimental Results}
\label{sec:R}

We evaluate the performance of the proposed method on both datasets (Cincinnati and Connecticut), with three other standard machine-learning techniques: 

\begin{itemize}
    \item Conv. Long-Short-Term-Memory-RNN (ConvLSTM) \cite{xingjian2015convolutional},
    \item Attention-based Conv. Long-Short-Term-Memory RNNs (Att-ConvLSTM)
    \item Time-Dist. Conv. Encoder-Decoder (TD-Conv-Enc-Dec)
\end{itemize}

As these architectures have different input configuration specifications, the input stacks of heat-maps are configured specific to each model. For Convolutional-LSTMs and TD-Conv-Enc-Dec, the input stack of heat-maps are scaled in the range [0,1]. For our approach, we utilize \textit{Wasserstein} distance as the adversarial loss function. To ensure stability in the training process, the input stacks are scaled within range [-1,1]. 

The layers used in building the Attention-based Conv-LSTM model \textit{(Att-ConvLSTM)} are documented sequentially as follows: 
64 $3\times3$ Conv-LSTMS with stride $1\times1$, max pooling with stride $2\times2$, 128 $3\times3$ Conv-LSTMS with stride $1\times1$, max pooling with stride $2\times2$, 256 $3\times3$ Conv-LSTMS with stride $1\times1$. To highlight the degree of importance that the features from each time-frame exhibit, we weigh each feature map using a softmax attention layer. Scaled Exponential Linear Unit (SELU) is used as the activation function in each convolutional block. 

The learning task can also be modeled on an encoder-decoder framework \textit{TD-Conv-End-Dec}, that handles data of different temporal scales. The architecture should be able to bridge the semantic gap between feature maps generated from temporal data.  Ronneberger et al. \cite{ronneberger2015u} and Drozdzal et al. \cite{drozdzal2016importance} systematically investigated the importance of skip connections to capture semantic links between feature maps. We build our model on \textit{UNet++}, an encoder-decoder architecture with nested skip-pathways, proposed by Zhou et. al. As our input data is a stack of time-dependent images, we use a series of nested time-distributed dense convolutional blocks. The nested skip pathways over the time-distributed convolutional layers aid in reducing the semantic gap between feature maps of the encoder and decoder, prior to feature-fusion. This is followed by aggregation of feature-information from $n$ different temporal-scales ($n=6$), where we employ global average-pooling over the flattened output of the last convolutional layer. Feature-reshaping is performed over the output of the global-pooling layer to generate the heat map.

\begin{figure*}
\includegraphics[width=0.75\textwidth, height = 0.3\textwidth]{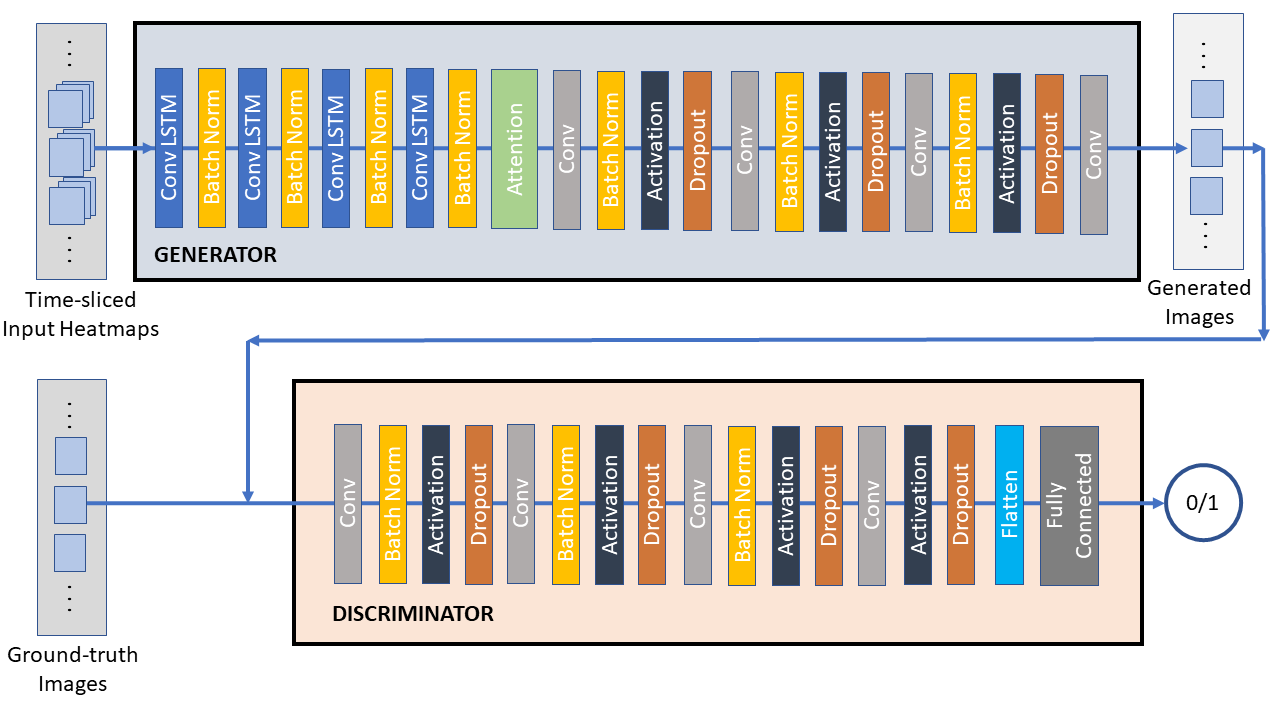}
\caption{Schematic Diagram of our GAN model}
\label{GAN}
\end{figure*}

\begin{figure*}
\includegraphics[width=0.75\textwidth]{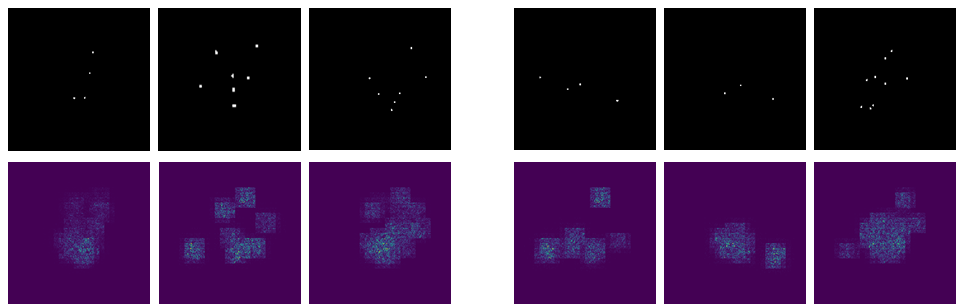}
\caption{Opioid Incidence Predictions for the Cincinnati (left) and Connecticut (right) datasets.}
\label{Results}
\end{figure*}

For our proposed model, \textit{`RMSProp'} is used as the model optimizer. Learning rate set to 0.00005. Our model is trained for 1500 epochs with a batch size of 8. The schematic diagram is given in Fig. \ref{GAN}. Mean-squared (MSE) and mean-absolute errors (MAE) are used as evaluation metrics for comparing the above-mentioned techniques with our proposed model. The results are presented in Tables \ref{table1}, \ref{table2}. Three daily Opioid incidence predictions for the Cincinnati and Connecticut datasets are illustrated in Fig. \ref{Results} respectively. The top row indicates the ground truth images for the respective daily incidences. The bottom row images are the predicted heat maps for the corresponding days. Greater the intensity, greater the likelihood of Opioid incidences occurring in that geographical area. 

\begin{table}[htbp!]
\centering
\caption{Evaluation results of the following methods on Cincinnati dataset.}
\setlength{\tabcolsep}{8pt}
\renewcommand{\arraystretch}{1.3}
\begin{tabular}{l|c|c}
\hline
\hline               
          \textbf{Method}  & \textbf{MSE} & \textbf{MAE} \\ 
\hline
\hline   
ConvLSTM         & 0.0628         & 0.0083       \\
Att-ConvLSTM    & 0.04256                  & 0.0062             \\  
TD-Conv-End-Dec  & 0.0562                 & 0.0075       \\ 
\textbf{Our Approach}  & \textbf{0.03371}   & \textbf{0.0057}           \\
\hline
\end{tabular}
\label{table1}
\vspace{-12.0pt}
\end{table}

\begin{table}[htbp!]
\centering
\caption{Evaluation results of the following methods on Connecticut dataset.}
\setlength{\tabcolsep}{8pt}
\renewcommand{\arraystretch}{1.3}
\begin{tabular}{l|c|c}
\hline
\hline               
          \textbf{Method}  & \textbf{MSE} & \textbf{MAE} \\ 
\hline
\hline   
ConvLSTM         & 0.0583         &   0.0081     \\
Att-ConvLSTM    & 0.0415                  & 0.0067              \\  
TD-Conv-End-Dec  & 0.0517                 & 0.0073       \\ 
\textbf{Our Approach}  & \textbf{0.0309}   & \textbf{0.0052}           \\
\hline
\end{tabular}
\label{table2}
\vspace{-8.0pt}
\end{table}

\section{Conclusion}
\label{sec:C}

In this work-in-progress paper, we have presented a novel deep learning architecture for generation of future heat maps, by analyzing past Opioid incidences, for the Cincinnati and Connecticut datasets. We have shown that our model arrives at accurate predictions even with a considerably small dataset. Due to the unavailability of such models in the literature that cater towards solving the proposed problem, the evaluation conducted in this paper are preliminary. However, they are substantial enough to highlight the significance of the problem and the efficacy of our approach.

\bibliographystyle{ACM-Reference-Format}
\bibliography{ACM-Reference-Format}
\end{document}